\documentclass[journal]{IEEEtran}
\usepackage {amsfonts}
\usepackage{bm}
\usepackage{chngcntr}
\usepackage{graphicx}
\usepackage{amsmath}
\usepackage{amsfonts}
\usepackage{amssymb}
\usepackage{latexsym}
\usepackage{float}
\usepackage{booktabs}

\usepackage{rotating}
\usepackage{multirow} 

\newtheorem{theorem}{Theorem}

\newtheorem{algorithm}[theorem]{Algorithm}

\newcommand{\E}{\mathbb{E}}

\newcommand{\bi}[1]{\mbox{\boldmath{$ #1 $}}} 
\counterwithin{equation}{section}
\usepackage{epstopdf}
\usepackage[final]{epsfig}

\begin{document}
%

\title{Robust and Efficient Boosting Method using the Conditional Risk}
%
%
%
%
%
%

\author{Zhi~Xiao, Zhe~Luo, Bo~Zhong,      
        and~Xin~Dang$^*$,~\IEEEmembership{Member,~IEEE}
\thanks{Z. Xiao is with the Department
of Information Management, Chongqing University, China.
E-mail: xiaozhi@cqu.edu.cn.}
\thanks{Z. Luo is with Bank of China, 151 National Road, Qingxiu District, Nanning, China. E-mail: shuilifang\_1988@126.com.}
\thanks{B. Zhong is with the Department
of Statistics and Actuarial Science, Chongqing University, China.
E-mail: zhongbo@cqu.edu.cn.}
\thanks{X. Dang$^*$ is the corresponding author and she is with the Department
of Mathematics, University of Mississippi, University, MS, 38677, USA.
E-mail: xdang@olemiss.edu.}}
\maketitle

\begin{abstract}

Well-known for its simplicity and effectiveness in classification, AdaBoost, however, suffers from overfitting when class-conditional distributions have significant overlap. Moreover, it is very sensitive to noise that appears in the labels. This article tackles the above limitations simultaneously via optimizing a modified loss function (i.e., the conditional risk). The proposed approach has the following two advantages. (1) It is able to directly take into account label uncertainty with an associated label confidence. (2) It introduces a ``trustworthiness" measure on training samples via the Bayesian risk rule, and hence the resulting classifier tends to have finite sample performance that is superior to that of the original AdaBoost when there is a large overlap between class conditional distributions. Theoretical properties of the proposed method are investigated. Extensive experimental results using synthetic data and real-world data sets from UCI machine learning repository are provided. The empirical study shows the high competitiveness of the proposed method in predication accuracy and robustness when compared with the original AdaBoost and several existing robust AdaBoost algorithms. 

\end{abstract}


%
\IEEEpeerreviewmaketitle

\section{Introduction}
For classification, AdaBoost is well-known as a simple but effective boosting algorithm with the goal of constructing a strong classifier by gradually combining weak learners \cite{Yoav97, Freund96, Schapire12}.  Its improvement on  classification accuracy benefits from the ability of adaptively sampling instances for each base classifier in the training process, more specifically in its re-weighting mechanism. It emphasizes the instances that were previously misclassified, and it decreases the importance of those that have been adequately trained. This adaptive scheme, however, causes an overfitting problem for noise data or data from overlapping class distributions \cite{Dietterich00, Melville04, Vezhnevets07}. The problem stems from the uncertainty of observed labels. It is usually a great challenge to do classification for the cases with overlapping classes. Also, it is both expensive and difficult to obtain reliable labels \cite{Frenay14}. In some applications (such as  biomedical data), perfect training labels are almost impossible to obtain. Hence, how to make AdaBoost achieve noise robustness and avoid overfits becomes an important task.  The aim of this paper is to construct a modified AdaBoost classification algorithm with a new perspective for tackling those problems. 

\subsection{Related Work}
Modifications to AdaBoost in dealing with noise data can be summarized into three strategic categories.  The first one introduces some robust loss functions as new criteria to be minimized, rather than using the original exponential loss.  The second type focuses on modifying the re-weighting rule in iterations in order to reduce or eliminate the effects of noisy data or outliers in the training sets.  The third approach suggests  more modest methods to combine weak learners that take advantage of base classifiers in other ways.

LogitBoost \cite{Friedman00} is an outstanding example of a modification of the first strategic category.  It uses the negative binomial log-likelihood loss function, which puts relatively less influence on instances with large negative margins\footnote{Margin is generally defined as $yf(\bi x)$, a negative margin implies a misclassification on an instance} in comparison with the exponential loss, thus LogitBoost is less affected by contaminated data \cite{Hastie01}. Based on the concept of robust statistics, Kanamori et al. \cite{Kanamori07} studied loss functions for robust boosting and proposed a transformation of loss functions in order to construct boosting algorithms more robust against outliers. Their usefulness has been confirmed empirically. However, the loss function they utilized was derived without considering efficiency. Onoda \cite{Onoda07} proposed a set of algorithms  that incorporate a normalization term into the original objective function to prevent from overfitting.  Sun et al. \cite{Sun04} and Sun et al. \cite{Sun06} modified AdaBoost using the regularization method. The approaches of the first category modification mainly differ in the loss functions and optimization techniques that are used. Sometimes, in the pursuit of robustness, it is hard to balance the complexity of a loss function with its computation cost. 

In general, modification of a loss function leads to a new re-weighting rule for AdaBoost, but some heuristic algorithms directly rebuild their weight updating scheme to avoid skewed distributions of examples in the training set. For instance, Domingo and Watanabe \cite{Domingo00} proposed MadaBoost that bounds the weight assigned to every sample by its initial probability. Zhang et al. \cite{Zhang08b} introduced a parameter into the weight updating formula to reduce weight changes in the training process. Servedio \cite{Servedio03} provided a new boosting algorithm, SmoothBoost, which produces only smooth distributions of weights but enables generation of a large margin in the final hypothesis. Utkin and Zhuk \cite{Utkin14} took the minimax (pessimistic) approach to search the optimal weights at each iteration in order to avoid outliers being heavily sampled in the next iteration.   

Since the ensemble classifier in AdaBoost predicts a new instance by a weighted majority voting among weak learners, the classifier that achieves high training accuracy will greatly impact the predictive result because of its large coefficient. This can have a detrimental effect on the generalization error, especially when the training set itself is corrupted \cite{Schapire99, Allende07}.  With this in mind, the third strategy seeks to provide a better way to combine weak learners.  Schapire and Singer \cite{Schapire99} improved boosting in an extended framework where each weak hypothesis produces not only classifications but also confidence scores in order to smooth the predictions. Besides, another method called Modest AdaBoost \cite{Vezhevets05} intends to decrease contributions of base learners in a modest way and forces them to work only in their domain.

The algorithms described above mainly focus on some robustifying principle, but they do not consider specific information in the training samples. Many other researches \cite{Takenouchi04, Kanamori04, Hayashi12} introduced the noise level into the loss function and extended some of the above mentioned methods. Nevertheless, most of these algorithms do not fundamentally change the fact that misclassified samples are weighted more than they are in the previous stage, though the increment of weights is smaller than that in AdaBoost. Thus mislabeled data may still hurt the final decision and cause overfitting.

In recent studies, many researchers were inclined to utilize the instance-base method to make AdaBoost robust against label noise or outliers. They evaluated the reliability or usefulness of each sample using statistical methods, and took that information into account.  Cao et al. \cite{Cao12} suggested a noise-detection based loss function that teaches AdaBoost to classify each sample into a mostly agreed class rather than using its observed label.  Gao and Gao \cite{Gao10} set the weight of suspicious samples in each iteration to zero and eliminated their effects in AdaBoost. Essentially, these two methods use dynamic correcting and deleting techniques in the training process. In \cite{Vezhnevets07}, the boosting algorithm directly works on a reduced training set whose ``confusing" samples have been removed. Zhang and Zhang \cite{Zhang08a} considered a local boosting algorithm.  Its reweighting rule and the combination of multiple classifiers utilize more local information of the training instances.

For handling label noise, it is natural to delete or correct suspicious instances first and then take the remaining ``good" samples as prototypes for learning tasks. This idea is not just for AdaBoost but is also applicable to general methods in many fields (e.g., \cite{Thongkam09}). Some approaches aim at constructing a good noise purification mechanism under the framework of different methods, such as ensemble methods \cite{Verbaeten03, Brodley96, Brodley11}, KNN or its variants \cite{Sanchez03, Liu12, Jiang04} and so on. Data preprocessing technique is a necessary step to improve quality of the prediction models in some cases \cite{Saez12}. However, some correct samples along with some valuable information may be discarded, and in the meantime, some noise samples may be included or some new noise samples may be introduced.  This is the limitation of correcting and deleting techniques.  To overcome this weakness,  Rebbapragada and Brodley \cite{Rebbapragada07} tried to use the confidence on the observed label as a weight of each instance during the training process and provided a novel framework for mitigating class noises. They showed empirically that this confidence weighting approach can outperform the discarding approach, but this new method was only applied to tree-based C4.5 classifier. The confidence-labeling technique they utilized fails to be a desirable label correction method. In \cite{Wang13} and \cite{Zhou14}, they considered and estimated the probability of an instance being from class 1 and used it as a soft label of the instance. 

\subsection{An overview of the proposed approach}
Inspired by instance-base methods and construction of robust algorithms, we propose a novel boosting algorithm based on label confidence, called CB-AdaBoost.  The observed label of each instance is treated as uncertain.  Not only the correctness, but also the degree of correctness of the label,  are evaluated according to a certain criterion before the training procedure.  We introduce the confidence of each instance into the exponential loss function. With such a modification, the misclassified and correctly classified exponential losses are weightily averaged. The weights are their corresponding probabilities represented by the correctness certainty parameter. In this way, the algorithm treats instances differently based on their confidence, and thus, it moderately controls the training intensity for each observation. The modified loss function is indeed the conditional risk or inner risk, which is quite different from a asymmetric loss or fuzzy loss.

Our method can make a smooth transition between full acceptance and full rejection of a sample label, thereby achieving robustness and efficiency at the same time.  In addition, our label-confidence based learning has no threshold parameter, whereas correcting and deleting techniques have to define a confidence level for ``suspect" instances so that they are relabeled or discarded in the training procedure. We derive theoretical results and also provide empirical evidences to show superior performance of the proposed CB-AdaBoost.

The contributions of this paper are as follows.
\begin{itemize}
\item A new loss function. We consider the conditional risk so that label uncertainty can be directly dealt with by the concept of label confidence.  This new loss function also leads the consideration of the sign of Bayesian risk rule on each of the sample points at the initialization of the procedure. 
\item A simple modification of adaptive boosting algorithm. Based on the new exponential loss function, AdaBoost has a simple explicit optimization solution at each iteration. 
\item Theoretical and empirical justifications for efficiency and robustness of the proposed method.
\item Consistency of the CB-AdaBoost is studied. 
\item Broad adaptivity. The proposed CB-AdaBoost is suitable for noise data and for class-overlapping data.  
\end{itemize}

\subsection{Outline of the paper}

 The remainder of the paper is organized as follows. Section \ref{sec:adaboost} reviews the original AdaBoost. In Section \ref{sec:method}, we propose a new AdaBoost algorithm.  We discuss in detail assignment of label confidence, the loss function, and the algorithm as well as its ability of adaptive learning in the label-confidence framework. Section \ref{sec:consistency} devotes to a study of the consistency property.  In Section \ref{sec:simulation}, we illustrate how the proposed algorithm works and investigate its performance through empirical studies of both synthetic and real-world data sets. Finally, the paper concludes with some final remarks in Section \ref{sec:conclusion}. A proof of consistency is provided in Appendix.

\section{Review of AdaBoost Algorithm}\label{sec:adaboost}

For binary classification, the main idea of AdaBoost is to produce a strong classifier by combining weak learners. This is obtained through an optimization that minimizes the exponential loss criterion over the training set. Let ${\cal  L}=\{(\bi x_i,z_i )_{i=1}^n\}$ denote a given training set consisting of $n$ independent training observations, where $\bi x_i=(x_{i1},x_{i2}, \cdots, x_{ip} )^T \in {\mathbb R}^p$ and $z_i \in \{1,-1\}$ represent the input attributes and the class label of the $i^{th}$ instance, respectively. The pseudo-code of AdaBoost is given in Algorithm \ref{Alg:AdaBoost} below.

\vskip 0.15in
\begin{algorithm}\label{Alg:AdaBoost}{AdaBoost Algorithm}
\begin{tabbing}
\texttt{Input}: \=${\cal L}= \{(\bi x_i,z_i)_{i=1}^n\}$ and the maximum number \\
\>of base classifiers $M$. \\
{\tt Initialize}: \= For $\forall i $, $w_i^{(1)}=1/N$, $D_i^{(1)}=w_i^{(1)}/S_1$,\\
\> where $S_1=\sum_{i=1}^n w_i^{(1)}$ is the \\
\>normalization factor. \\
{\tt For} \= $m=1$ {\tt To} $M$\\
\> \texttt{1} \= Draw instance from ${\cal L}$ with replacement \\
\>\>according to  the distribution $D_i^{(m)}$ to form\\
\>\> a training set ${\cal L}_m$;\\
\> \texttt{2} \= Train ${\cal L}_m$ with the base learning algorithm\\ 
\>\> and obtain a weak hypothesis $h_m$; \\
\> \texttt{3} \= Compute $\varepsilon_m=\sum_{i: h_m(\bi x_i)\neq z_i}^n D_i^{(m)}$; \\
\> \texttt{4} \= Let $\beta_m =\frac{1}{2}\ln \left(\frac{1-\varepsilon_m}{\varepsilon_m}\right)$;\\
\>\> {\tt If}  $\beta_m<0$, {\tt then} $M=m-1$ and {\tt abort} \\
\>\>{\tt loop}.\\
\>\texttt{5} \= Update $w_i^{(m+1)}=w_i^{(m)}e^{-z_i\beta_mh_m(\bi x_i)}$; \\
\>\>$D_i^{(m+1)}=w_i^{(m+1)}/S_{m+1}$ for $\forall i$, where \\
\>\>$S_{m+1}=\sum_{i=1}^n w_i ^{(m+1)}$;\\
{\tt End For}\\
\texttt{Output}: sign($\sum_{m=1}^M\beta_m h_m(\bi x)$).
\end{tabbing}
\end{algorithm}
\vskip 0.15in

In the AdaBoost Algorithm, the current classifier $h_m$ is induced on the weighted sampling data,  and the resulting weighted error $\varepsilon_m$ is computed. The individual weight of each of the observations is updated for the next iteration. AdaBoost is designed for clean training data---that is, each label $z_i$ is the true label of $\bi x_i$.  In this framework, any instance was previously misclassified has a higher probability to be sampled in the next stage.  In this way, the next classifier focuses more on those misclassified instances,  and hence, the final ensemble classifier achieves high accuracy. For mislabeled data, however, those observations which were misclassified in the previous step are weighted less,  and those correctly classified instances are weighted more  than they should. This leads to the next training set ${\cal L}_{m+1}$ being seriously corrupted, and those mislabeled data eventually hurt the performance of the ensemble classifier. Therefore, some modifications should be introduced to make AdaBoost insensitive to class noise. 

\section{Label-confidence based Boosting Algorithm}
\label{sec:method}

\subsection{Label confidence}
For the class noise data problem, the observed label $y$ associated with $\bi x$ may be incorrectly assumed due to some random mechanism.  For the class overlapping problem, the label $y$ associated with $\bi x$ is a realization of random label from some distribution.  In our approach to deal with both problems, we treat the true label $Z$ to be random. Let $y$ (either 1 or -1) be the observed label associated with $\bi x$. We define a parameter $\gamma$ as the probability of being correctly labeled, that is, $\gamma=P(Z=y|\bi x)$ and $P(Z=-y|\bi x)=1-\gamma$ for $\gamma\in [0,1]$.  The quantity $|\gamma-(1-\gamma)|= (2\gamma-1) \mbox{sign}(2\gamma-1)$ measures ``trustworthiness" of label $y$ and $\mbox{sign}(2\gamma-1)=\pm 1$ represents confidence towards correctness or wrongness of the label.   Thus we can use $\mbox{sign}(2\gamma-1)y$ as the trusted label with confidence level $|2\gamma-1|$. For example, for $\gamma=1$, $|2\gamma-1|=1$ and $\mbox{sign}(2\gamma-1)=1$ represent that we are $100\%$ confident about correctness of the label $y$, while for $\gamma=0$, $|2\gamma-1|=1$ and $\mbox{sign}(2\gamma-1)=-1$ represent $100\%$ certainty  about the wrongness of $y$ so that $-y$ should be $100\%$ trusted. The label $y$ with $\gamma=0.5$ is the most unsure or fuzzy case with 0 confidence.  It is easy to see that the trusted label $\mbox{sign}(2\gamma-1)y$ is exactly the Bayes rule. Let $\eta(\bi  x) = P(Z=1|\bi x)$ and hence the Bayes rule is $\mbox{sign}(2 \eta(\bi x)-1)$, which is equal to $\mbox{sign}(2\gamma-1)y$ for both $y=1$ and $y=-1$. 

For given training data ${\cal L}=\{(\bi x_1,y_1),...,(\bi x_n,y_n)\}$, let a parameter vector $\bi \gamma=(\gamma_1,\gamma_2,...,\gamma_n )$ represent their probabilities of being correctly labeled. That is, the parameter $\gamma_i$ can be regarded as the confidence of a sample $\bi x_i$ being correctly labeled as $y_i$.  In the next subsections, we first introduce the modified loss function based on a given $\bi \gamma$, then propose the confidence based adaptive boosting method (CB-AdaBoost).  At the end of the section we discuss the estimation of $\bi \gamma$.

\subsection{Conditional-risk loss function}
Given a clean training set with correct labels $z_i$'s available, the original AdaBoost minimizes the empirical exponential risk
\begin{equation}\label{eqn:emprisk}
\hat{risk}(f) =\frac{1}{n} \sum_{i=1}^n \exp(-z_i f(\bi x_i))
\end{equation}
over all linear combinations of base classifiers in the given space ${\cal H}$,  assuming that an exhaustive weak learner returns the best weak hypothesis on every round \cite{Friedman00, Schapire12}. Now in class noise data, the true label $z_i$ is unknown. We only observe $y_i$ associated with $\bi x_i$. Based on the assumption, given $\bi x_i$, the probability that $Z_i$ is $y_i$ is $\gamma_i$. It is natural to consider the following empirical risk:
\begin{equation}\label{eqn:risk}
\hat{R}=\frac{1}{n} \sum_{i=1}^n [\gamma_i \exp(-y_if(\bi x_i) )+(1-\gamma_i)\exp(y_if(\bi x_i))].
\end{equation}
That is, we treat the observed label $y_i$ as a fuzzy label with $\gamma_i$ correctness confidence.  In other words, we consider the modified exponential loss function
\begin{equation}\label{eqn:loss}
L_{\gamma} (y,f(\bi x))=\gamma \exp(-yf(\bi x))+(1-\gamma)\exp(yf(\bi x)), 
\end{equation}
which has a straightforward interpretation. The label $y$ associated with $\bi x$ is trusted with $\gamma$ confidence and it is corrected as $-y$ with $1-\gamma$ confidence.  It is easy to check that  the loss (\ref{eqn:loss}) 
$$L_{\gamma}(y,f(\bi x)) = E_{z|\bi x} \exp(-zf(\bi x)),$$ 
which is the inner risk defined in \cite{Steinwart08}. The reason it is called the inner risk is because the true exponential risk is 
\begin{eqnarray}
risk(f) &=& \E \exp(-zf(\bi x))  \label{jointrisk}\\
&=&\E_{\bi x}\E_{z|\bi x}[\exp(-zf(\bi x)]  \nonumber\\
&=&\E_{\bi x} L_{\gamma}(y,f(\bi x)) \label{inner}
\end{eqnarray}
for $y=\pm 1$.  From this perspective, we consider minimizing the empirical inner  risk of (\ref{inner}), while the original AdaBoost minimizes the empirical risk of (\ref{jointrisk}). Steinwart and Christmann \cite{Steinwart08} showed in their Lemma 3.4 that the risk can be achieved by minimizing the inner risks, where the expectation is taken with respect to the marginal distribution of $\bi x$, in contrast to (\ref{jointrisk}) where the expectation is taken with respect to the joint distribution of $(\bi x, z)$.  Clearly, under the scenarios of overlapping class and label noise, the empirical inner risk (\ref{eqn:risk}) has an advantage over (\ref{eqn:emprisk}). 

In \cite{Bartlett06}, (\ref{eqn:loss}) is called the conditional $\psi$-risk with $\psi$ being the exponential loss function. A classification-calibrated condition on the conditional risk is provided to ensure a pointwise form of Fisher consistency for classification. In other words, if the condition is satisfied, the 0-1 loss can be surrogated by the convex $\psi$ loss in order to make the minimization computationally efficient.  The exponential loss is classification-calibrated. Our proposed method utilizes a different empirical estimator of the exponential risk. Its consistency follows from the consistency result of AdaBoosting \cite{Bartlett07} along with consistent estimation of $\gamma$. More details are presented in Section \ref{sec:consistency}.
 
The loss (\ref{eqn:loss}) is closely related to the asymmetric loss used in the literature (e.g. \cite{Wang12, Masnadi11}),   but the motivation and goal of the two losses are quite different. The asymmetric loss treats two classes unequally. Two misclassification errors produce different costs.  However, the costs or weights  do not necessarily sum up to 1. In asymmetric loss, the ratio of two costs is usually used to measure the degree of asymmetry and  is often a constant parameter, while in (\ref{eqn:loss}) it is a function of $\bi x$. Also the loss  (\ref{eqn:loss}) takes a linear combination of the exponential loss at $y$ and $-y$, while the asymmetric loss only takes one. 

Indeed, $\gamma$ in the loss (\ref{eqn:loss}) is the posterior probability used in \cite{Tao05} for the support vector machine technique. The similarity is that we all use the sign of the Bayes rule as the trusted label. However, we also include the magnitude $|2\gamma-1|$ in our loss function. We associate the trusted label with a confidence $|2\gamma-1|$, while in \cite{Tao05} the confidence is always 1. The idea of label confidence is closely related to fuzzy label used in fuzzy support vector machines \cite{Lin02}. The difference is that fuzzy label only assigns an importance weight for the observed label without considering its correctness. 

Next, we derive the proposed method based on the modified exponential loss function.

\subsection{Derivation of our algorithm}
For an additive model,
\begin{equation}\label{eqn:boost}
f_M (\bi x)=\sum_{m=1}^M\beta_mh_m(\bi x),
\end{equation}
where $h_m(\bi x)\in \{-1,1\}$ is a weak classifier in the $m^{th}$ iteration, $\beta_m$  is its coefficient and $f_M (\bi x)$  is an ensemble classifier. Our goal is to learn an ensemble classifier with a forward stage-wise estimation procedure by fitting an additive model to minimize the modified loss functions. Let us consider an update from $f_{m-1} (\bi x)$ to $f_m(\bi x) =f_{m-1} (\bi x)+\beta_m h_m (\bi x)$ by minimizing (\ref{eqn:risk}). This is an optimization problem to find solutions $h_m$ and  $\beta_m$, that is,
\begin{align}
(\beta_m, h_m)&=\arg\min_{\beta,h} \sum_{i=1}^n[\gamma_i \exp({-y_i f_m(\bi x_i )}) \nonumber\\
&+(1-\gamma_i ) \exp({y_i f_m(\bi x_i ) }) ] \nonumber \\
&=\displaystyle \arg\min_{\beta,h} \sum_{i=1}^n [w_{1i}^{(m)} \exp(-y_i \beta h(\bi x_i)) \nonumber\\
&+w_{2i}^{(m) } \exp(y_i \beta h(\bi x_i ))],  \label{eqn:sum}
\end{align}
where $w_{1i}^{(m) }=\gamma_i e^{-y_i f_{m-1} (\bi x_i ) }$ and $w_{2i}^{(m) }=(1-\gamma_i ) e^{y_i f_{m-1} (\bi x_i ) }$ are independent with  $h_m$ and $\beta_m$. 

As we will show, $h_m$ and $\beta_m$ can be derived separately in two steps.  Let us first optimize the weak hypothesis $h_m$. The summation in (\ref{eqn:sum}) can be expressed alternatively as 
\[
\begin{array}{l}
\sum_{i=1}^n [w_{1i}^{(m)} \exp(-y_i \beta h(\bi x_i))+w_{2i}^{(m) } \exp(y_i \beta h(\bi x_i ))]\\\\
=\sum_{\{i: h(\bi x_i )=y_i\}}^n [w_{1i}^{(m) } e^{-\beta}+w_{2i}^{(m) } e^\beta] \\\\
+\sum_{\{i:h(\bi x_i )\neq y_i\}}^n[w_{1i}^{(m) } e^{\beta}+w_{2i}^{(m) } e^{-\beta} ]\\\\
=\sum_{i=1}^n [w_{1i}^{(m) } e^{-\beta}+w_{2i}^{(m) } e^{\beta} ] \\\\
+(e^{\beta} -e^{-\beta} ) \sum_{\{i: h(\bi x_i )\neq y_i\}}^N [w_{1i}^{(m) }-w_{2i}^{(m) } ].
\end{array}
\]
Therefore, for any given value of $\beta >0$,  (\ref{eqn:sum}) is equivalent to the minimization of
\begin{equation}\label{eqn:hm}
 h_m=\arg\min_h\sum_{i=1}^n [w_{1i}^{(m) }-w_{2i}^{(m) } ] I\{h(\bi x_i )\neq y_i \}.
\end{equation}
It is worthwhile to mention that the term $(w_{1i}^{(m) }-w_{2i}^{(m) } )$ may be negative, hence it  cannot be directly interpreted as the ``weight" of the instance $(x_i,y_i )$ in the training set.  According to the analytical solution of $h_m$, the base classifier is expected to correctly predict $(\bi x_i, y_i )$ in the case of $w_{1i}^{(m) } \geq w_{2i}^{(m) }$ and otherwise misclassify $(\bi x_i, y_i )$. This is equivalent to solving 
 \begin{equation}\label{eqn:hm2}
 \min_h\sum_{i=1}^n | w_{1i}^{(m) }-w_{2i}^{(m) }| I \left\{h(\bi x_i )\neq \mbox{sign}([w_{1i}^{(m) }-w_{2i}^{(m) }] y_i )\right\}.
 \end{equation}
In other words, $h_m$  is actually the one that minimizes the prediction error over the  set $\{(\bi x_i, \mbox{sign}([w_{1i}^{(m) }-w_{2i}^{(m) }] y_i )_{i=1}^n \}$ with each instance weighted $|w_{1i}^{(m) }-w_{2i}^{(m) }|$. In each iteration, we treat $\mbox{sign}([w_{1i}^{(m) }-w_{2i}^{(m) }] y_i)$ as the label of $\bi x_i$ and $|w_{1i}^{(m) }-w_{2i}^{(m) }|$ as its importance. This provides a theoretical justification of the sampling scheme in our proposed algorithm, which is given later.

Next, we optimize $\beta_m$. With $h_m$ fixed, $\beta_m$  minimizes
\begin{align}\label{eqn:bm}
&\sum_{i:h_m (\bi x_i )=y_i}[w_{1i}^{(m)} e^{-\beta}+w_{2i}^{(m)} e^\beta] \nonumber\\
& +\sum_{i: h_m (\bi x_i )\neq y_i}[w_{1i}^{(m) } e^\beta+w_{2i}^{(m) } e^{-\beta} ].
\end{align}
Upon setting the derivative of (\ref{eqn:bm}) (with respect to $\beta$) to zero, we obtain
\begin{equation} \label{eqn:bmd}
\beta_m=\frac{1}{2} \ln \frac{\sum_{i: h_m (\bi x_i )=y_i} w_{1i}^{(m) }+\sum_{i: h_m (\bi x_i )\neq y_i} w_{2i}^{(m)}} {\sum_{i: h_m (\bi x_i )\neq y_i} w_{1i}^{(m) }+\sum_{i: h_m (\bi x_i )= y_i} w_{2i}^{(m)}}.
\end{equation}
Note that the condition that
\begin{align} \label{eqn:cond}
&\sum_{i: h_m (\bi x_i )=y_i} w_{1i}^{(m) }+\sum_{i: h_m (\bi x_i )\neq y_i} w_{2i}^{(m)} \nonumber\\
&> {\sum_{i: h_m (\bi x_i )\neq y_i} w_{1i}^{(m) }+\sum_{i: h_m (\bi x_i )= y_i} w_{2i}^{(m)}}
\end{align}
should hold in order to ensure the value of $\beta_m$ is positive.

The approximation on the $m^{th}$ iteration is then updated as
\begin{equation}\label{eqn:miter}
f_m (\bi x)=f_{m-1} (\bi x)+\beta_m h_m (\bi x),
\end{equation}
which leads to the following update of $w_{1i}^{(m) }$ and $w_{2i}^{(m) }$: 
$$w_{1i}^{(m+1) }=w_{1i}^{(m) } e^{-y_i \beta_m h_m (\bi x_i ) }$$
and
$$w_{2i}^{(m+1) }=w_{2i}^{(m) } e^{y_i \beta_m h_m (\bi x_i ) }. $$
By repeating the procedure above, we can derive the iterative process for all rounds $m\geq 2$ until $m=M$ or the condition (\ref{eqn:cond}) is not satisfied. The initial values take $w_{1i}^{(1)}=\gamma_i$ and $w_{2i}^{(1)}=1-\gamma_i$. Now we write the procedure into the pseudo-code of the Algorithm \ref{Alg:CBboost}.

\vskip 0.15in
\begin{algorithm}\label{Alg:CBboost}{CB-AdaBoost Algorithm}
\begin{tabbing}
\texttt{Input}: \= ${\cal L} =\{(\bi x_i,y_i )_{i=1}^n\}$, $\bi \gamma=\{(\gamma_i)_{i=1}^n\}$ and $M$ \\

{\tt Initialize}: \= For $\forall i $, $w_{1i}^{(1)}=\gamma_i$, $w_{2i}^{(1)}= 1-\gamma_i$,\\ \>$D_i^{(1)}=|w_{1i}^{(1)}-w_{2i}^{(1)}|/S_1$, where\\
\> $S_1=\sum_{i=1}^n |w_{1i}^{(1)}-w_{2i}^{(1)}|$ \\
{\tt For} \= $m=1$ {\tt To} $M$\\
\> \texttt{1} \= Relabel all instances in ${\cal L}$ to compose a new\\ 
\>\>data set as ${\cal L}'=\{(\bi x_i,y_i' )_{i=1}^n\}$, where $\bi x_i\in {\cal L}$, \\
\>\>$y_i'=\mbox{sign}[(w_{1i}^{(m)}-w_{2i}^{(m)}) y_i ]$;\\
\> \texttt{2} \= Draw instance from ${\cal L}'$ with replacement\\ 
\>\>according to the distribution $D_i^{(m)}$ to \\
\>\>compose a training set ${\cal L}_m$;\\
\> \texttt{3} \= Train ${\cal L}_m$ with the base learning algorithm\\ \>\> and obtain
a weak hypothesis $h_m$; \\

\> \texttt{4} \= Let $\beta_m=\frac{1}{2} \ln \frac{\sum_{i: h_m (\bi x_i )=y_i} w_{1i}^{(m) }+\sum_{i: h_m (\bi x_i )\neq y_i} w_{2i}^{(m)}} {\sum_{i: h_m (\bi x_i )\neq y_i} w_{1i}^{(m) }+\sum_{i: h_m (\bi x_i )= y_i} w_{2i}^{(m)}}$ \\
\>\> {\tt If}  $\beta_m<0$, {\tt then} $M=m-1$ and {\tt abort} \\ 
\>\>{\tt loop}.\\
\>\texttt{5} \= Update $w_{1i}^{(m+1)}=w_{1i}^{(m)}e^{-y_i\beta_mh_m(\bi x_i)}$; \\
\>\> $w_{2i}^{(m+1)}=w_{2i}^{(m)}e^{y_i\beta_mh_m(\bi x_i)}$;\\
\>\>$D_i^{(m+1)}=|w_{1i}^{(m+1)}-w_{2i}^{(m+1)}|/S_{m+1}$ for $\forall i$, \\
\>\> where $S_{m+1}=\sum_{i=1}^n|w_{1i}^{(m+1)}-w_{2i}^{(m+1)}|$;\\
{\tt End For}\\
\texttt{Output}: sign($\sum_{m=1}^M\beta_m h_m(\bi x)$).

\end{tabbing}
\end{algorithm}
\vskip 0.15in

\subsection{Class noise mitigation}
In this subsection, we study the effect of label confidence, and we investigate the adaptive ability of CB-AdaBoost in the mitigation of overfitting and class noise from aspects of its re-weighting procedure and classifier combination rule.

First, the initialization of distribution shows different initial emphases on training instances between Algorithm \ref{Alg:AdaBoost} and Algorithm \ref{Alg:CBboost}. As discussed early, $|\gamma_i-(1-\gamma_i)|$ actually represents its label certainty, and it is used as the initial weight in Algorithm \ref{Alg:CBboost}.  The conditional risk type of loss function leads this initialization and the weighting strategy that distinguishes instances based on their own confidences. Consequently, the instances with a high certainty receive a priority to be trained. This makes sense as these instances are usually those identifiable from a statistical standpoint, and thus, they are more valuable in classification. By contrast, Algorithm \ref{Alg:AdaBoost} treats each instance equally at the beginning without considering the reliability on the samples.

Second,  we consider $y_i'=\mbox{sign}(2\gamma_i-1)y_i$ as the label of $\bi x_i$ in Algorithm \ref{Alg:CBboost}. Under the mislabeled or class overlapping scenarios, this design makes sense because $\mbox{sign}(2\gamma_i-1)$ represents the confidence towards correctness or wrongness of the label $y_i$. If $\mbox{sign}(2\gamma_i-1)=1$, $y_i$ should be trusted with confidence $|2\gamma_i-1|$. Nevertheless, if $\mbox{sign}(2\gamma_i-1)=-1$, $-y_i$ should be trusted with confidence $|2\gamma_i-1|$.  The original AdaBoost trusts label $y_i$ completely, which is inappropriate under mislabelling and class overlapping. As shown before, the trust label $y_i'$ in CB-AdaBoost has the same sign as the Bayes rule at sample point $\bi x_i$. Intuitively, our method takes more information at the initialization.

Third, we take a detailed look at the weight updating formulas in Algorithm \ref{Alg:CBboost} and subsequently obtain the following results on the first re-weighting process. We say that an instance $\bi x_i$ is misclassified at the $m^{th}$ iteration if $h_m(\bi x_i)\neq y_i'$, where $y_i' = \mbox{sign}[(w_{1i}^{m}-w_{2i}^{m}) y_i]$; otherwise, it is correctly classified. \\[1ex]

\noindent{\bf Proposition 1}. {\em The misclassified instance receives larger weight for the next iteration.}\\[1ex]

\noindent{\bf Proof}. Two types of misclassification are either $h_m (x_i )\neq y_i$ with $w_{1i}^{(m)}>w_{2i}^{(m) }$ or $h_m(x_i) \neq -y_i$ with $w_{1i}^{(m)}<w_{2i}^{(m) }$. In the first case, 
\begin{align*}
&|w_{1i}^{(m+1) }-w_{2i}^{(m+1) } |=|w_{1i}^{(m) } e^{\beta_m}-w_{2i}^{(m)} e^{-\beta_m } |\\
&>|w_{1i}^{(m)}-w_{2i}^{(m)} |,
\end{align*}
while in the second case, 
\begin{align*}
&|w_{1i}^{(m+1) }-w_{2i}^{(m+1) } |=|w_{1i}^{(m) } e^{-\beta_m}-w_{2i}^{(m)} e^{\beta_m } |\\
&>|w_{1i}^{(m)}-w_{2i}^{(m)} |.
\end{align*}
In both cases, the weight increases.\\[1ex]

\noindent{\bf Proposition 2}. {\em If an instance is correctly classified and its certainty is high enough so that $\max\{w_{1i}^{(m)},w_{2i}^{(m)} \}>e^{\beta_m}\min\{w_{1i}^{(m)},w_{2i}^{(m)}\}$, then it receives smaller weight at the next iteration. }\\[1ex]

\noindent{\bf Proof}. We can easily check two cases. For the case of $w_{1i}^{(m)}>w_{2i}^{(m)}$ and $h_m (x_i )=y_i$, when $w_{1i}^{(m)} >e^{\beta_m}w_{2i}^{(m)}$, 
we have 
\begin{align*}
&|w_{1i}^{(m+1)}-w_{2i}^{(m+1)} |=|w_{1i}^{(m)} e^{-\beta_m}-w_{2i}^{(m)} e^{\beta_m} |\\
&<|w_{1i}^{(m)}-w_{2i}^{(m)} |. 
\end{align*}

For the case of $w_{1i}^{(m)}<w_{2i}^{(m)}$   and $h_m (x_i )=-y_i$, if $w_{1i}^{(m)} >e^{\beta_m}w_{2i}^{(m)}$, we have
\begin{align*}
&|w_{1i}^{(m+1)}-w_{2i}^{(m+1)} |=|w_{1i}^{(m)} e^{\beta_m}-w_{2i}^{(m)} e^{-\beta_m} |\\
&<|w_{1i}^{(m)}-w_{2i}^{(m)}|.
\end{align*}

Propositions 1 and 2 show that on the first important stage, CB-AdaBoost inherits the adaptive learning ability of AdaBoost and has the distinction that it adjusts the distribution of instances according to the current classification with respect to the commonly agreed information. Moreover, the degree of adjustment is managed by the confidence of each sample. For the following iterations, we can imagine the resampling process.  The weights of instances with high confidence stay at a high level until most of them are sufficiently learned. After that, their proportion decreases rapidly while the proportion of instances with low confidence increases gradually. As uncertain instances consist of most of the training set, the training process is difficult to continue. On the other hand, once a new classifier becomes no better than a random guess, then an early stop in the iterative process is possible. This is because the condition (\ref{eqn:cond}) no longer holds in that case.  Thus, our proposed method effectively prevents the ensemble classifier from overfitting.

Fourth, let us scrutinize the classifier ensemble rule. \\[1ex]

\noindent{\bf Proposition 3}. {\em In the framework of Algorithm \ref{Alg:CBboost}, define $\varepsilon_m'$ as the error rate of $h_m$ over its training set ${\cal L}_m$ during the $m^{th}$ iteration---that is,  $\varepsilon_m' =\sum_{i:h_m (\bi x_i )\neq y_i'}^n |w_{1i}^{(m) }-w_{2i}^{(m)} |/S_m$. We then have
$$\beta_m<\frac{1}{2}\ln \left(\frac{1-\varepsilon_m'}{\varepsilon_m' }\right).$$}

\noindent {\bf Proof}. We can prove this result by giving an equivalent representation of $\beta_m$ as:
\begin{align*}
&\beta_m=\frac{1}{2}\ln \left(\frac{\sum_{i:h_m (\bi x_i )=y_i}^N w_{1i}^{(m)}+\sum_{i:h_m (\bi x_i )\neq y_i}^n w_{2i}^{(m)}}{\sum_{i:h_m (\bi x_i )\neq y_i}^n w_{1i}^{(m)}+\sum_{i:h_m (\bi x_i )= y_i}^n w_{2i}^{(m)}}\right)\\
&= \frac{1}{2}\ln \left(\frac{\sum_{i:h_m(\bi x_i)=y_i'}^n |w_{1i}^{(m)}-w_{2i}^{(m)}|+c}{\sum_{i:h_m(\bi x_i)\neq y_i'}^n |w_{1i}^{(m)}-w_{2i}^{(m)}|+c}\right),
\end{align*}
where 
$$c=\sum_{i:w_{1i}^{(m)}<w_{2i}^{(m)}}^n w_{1i}^{(m)}+\sum_{i:w_{1i}^{(m)}>w_{2i}^{(m)}}^n w_{2i}^{(m)}.$$

With the Condition (\ref{eqn:cond}) being satisfied, we obtain
$\sum_{i: h_m(\bi x_i)=y_i'}^n |w_{1i}^{(m)}-w_{2i}^{(m)}| >\sum_{i: h_m(\bi x_i)\neq y_i'}^n |w_{1i}^{(m)}-w_{2i}^{(m)}|$, which implies
\begin{align*}
&\frac{1-\varepsilon_m'}{\varepsilon_m'} =\frac{ \sum_{i:h_m(\bi x_i)=y_i'}^n |w_{1i}^{(m)}-w_{2i}^{(m)}|}{\sum_{i:h_m(\bi x_i)\neq y_i'}^n |w_{1i}^{(m)}-w_{2i}^{(m)}|} \\
&>\frac{ \sum_{i:h_m(\bi x_i)=y_i'}^n |w_{1i}^{(m)}-w_{2i}^{(m)}|+c}{\sum_{i:h_m(\bi x_i)\neq y_i'}^n |w_{1i}^{(m)}-w_{2i}^{(m)}|+c}.
\end{align*}
Thus, the proof of Proposition 3 is complete. 

It turns out that  $\beta_m$ calculated in our modified algorithm does not take into account the full value of the odd ratio for each hypothesis. In fact, it is smaller than that calculated in AdaBoost, so our algorithm combines base classifiers and updates instance weights modestly.  This effectively avoids the situation where some hypotheses dominated by substantial classification noise are exaggerated by their large coefficients in the final classifier. 

We have studied the CB-AdaBoost algorithm in detail and compared its advantages to the original one. Next, we discuss the remaining issue of how to estimate label confidence. 

\subsection{Assignment of label confidence}\label{sec:assignment}

In most cases, since it is difficult to track the data collection process and identify where corruptions will most likely occur, we evaluate the confidence on labels according to the statistical characteristics of the data itself. In this regard, \cite{Rebbapragada07} suggested a pair-wise expectation maximization method (PWEM) to compute confidence of labels. Cao et al \cite{Cao12} applied KNN to detect suspicions examples. However, a direct application of these methods may not be efficient for data sets whose noise level is high. We believe that a cleaner data set can make a better confidence estimation. Therefore, before confidence assignment, a noise filter shall be introduced to eliminate  very suspicious instances so that we are able to extract  more reliable statistical characteristics from the remaining data. 

\begin{table*}[thb]
\renewcommand{\arraystretch}{1.3}
\centering
\begin{tabular}{lllccc} \hline\hline
&	&& \multicolumn{3}{c}{Noise Level}  \\ \cline{4-6}
&&&		$10\%$&	$20\%$&	$30\%$\\ \hline
Normal	&$n=50$&Clean	&$0.8919\pm0.2068$&$	0.8693\pm0.2267$&	$0.8201\pm0.2519$\\
	&&Mislabeled&$	0.0581\pm 0.0616$&$	0.1795\pm0.2265$&	$0.4459\pm0.4018$\\[1ex]
&$n=500$&Clean	&$0.9172\pm0.2011$&$	0.8547\pm0.1978$&	$0.7145\pm0.1514$\\
	&&Mislabeled&$	0.0850\pm 0.1790$&$	0.1446\pm0.1843$&	$0.2742\pm0.1499$\\  \hline
Sine	&$n=50$&Clean	&$0.8551\pm0.2503$&$	0.8503\pm0.2720$&	$0.7142\pm0.3556$\\
	&&Mislabeled&$	0.1833\pm 0.2905$&$	0.3888\pm0.4047$&	$0.4661\pm0.3999$\\ [1ex]
 &$n=500$&Clean	&$0.8731\pm 0.2639$	 &$0.8543\pm0.2675$& 	$0.8451\pm 0.2844$\\
& &Mislabeled	&$0.2870\pm0.3832$&	$0.4142\pm0.4195$&	$0.4958\pm 0.4257$\\ \hline\hline \\
\end{tabular}
\caption{Average and standard deviation of the confidences for clean and mislabeled samples in two data sets with different noise levels.  \label{conflevel}}
\end{table*}

First, a noise filter scans over the original data set.  Using a similarity measure between instances to find a neighborhood of each instance, one can compute the agreement rate for its label from its neighbors.  The instances with an agreement rate below a certain threshold are eliminated.  The above process can be repeated several times since some suspect instances may be exposed later when their neighborhood changes. In our experiment, the threshold is set to 0.07 at the beginning with an increment of 0.07 in each subsequent round. The process is repeated three times, and the final cut-off value for the agreement rate is 0.21 so that the sample size doesn't decrease much.  In the mean time, distributional information of the sample is kept relatively intact.    Once a filtered data set, denoted as ${\cal L}_{red}$, is obtained, two methods can be used to compute label confidence. 

If the noise level $\varepsilon$ over the training labels is known or can be estimated, we can represent the frequency of observations with label $y$ as follows:
\begin{align*}
&P(Y=y)=P(Y=y, Z=y)+P(Y=y,Z=-y) \\
&=(1-\varepsilon)P(Z=y)+\varepsilon P(Z=-y),
\end{align*}
where the noise level $\varepsilon= P(Y=y|Z=-y) =P(Y=-y|Z=y)$. This representation explains two sources for the composition of label $y$: correctly labeled instances belonging to true class $y$ and mislabeled instances belonging to true class $-y$. Then $P(Z=y) = (P(Y=y)-\varepsilon)/(1-2\varepsilon)$,  and utilizing the Bayesian formula, we assess the confidence as follows:
\[
\begin{array}{l}
\displaystyle \gamma=P(Z=y|\bi x)=\frac{P(Z=y) f(\bi x|Z=y)}{f(\bi x)}\\
 \displaystyle=\frac{P(Z=y)f(\bi x|Z=y)}{f (\bi x|Z=y)P(Z=y)+f(\bi x |Z=-y)P(Z=-y) }\\\\
\displaystyle=\frac{(P(Y=y)-\varepsilon)f( \bi x|Z=y)}{(P(Y=y)-\varepsilon) f( \bi x| Z=y)+\varepsilon f(\bi x|Z=-y)}.
\end{array}
\]
With conditional distribution type known, $f(\bi x|Z=y)$ and $f( \bi x|Z=-y)$ can be estimated under ${\cal L}_{red}$ while $P(Y=y)$ is directly set to be the sample proportion of class $y$ in ${\cal L}$.

The second method doesn't need to assume the noise level. KNN is recalled to assign confidence on each label. Based on ${\cal L}_{red}$, the label agreement rate of each instance among its nearest neighbors can act as its confidence. So the confidence probability of an example $(\bi x,y)$ in ${\cal L}$ is computed as follows:
\begin{equation} \label{eqn:KNN}
P(Z=y| \bi x)=\frac{1}{K}  \sum_{j=1}^K \sum_{\bi x_j \in {\cal N}(\bi x)} I(y_j=y) ,
\end{equation}
where ${\cal N}(\bi x)$  represents the set containing $K$ nearest neighbors of $\bi x$ from ${\cal L}_{red}$. In our experiment, $K=5$ is used. 

In the simulation of Section \ref{sec:simulation}, we will evaluate the quality of confidence assigned by these two methods. In practice, however, the Bayesian method is usually infeasible since the noise level is unknown.

\subsection{Relationship to previous work}
Note that our modified algorithm reduces to AdaBoost  if we set the confidence on each label to one.   The greater the confidence on each instance, the less CB-AdaBoost differs from AdaBoost in terms of the weight updating and base classifiers, as well as their coefficients in successive iterations.
Rebbapragada et al. \cite{Rebbapragada07} proposed instance weighting via confidence in order to mitigate class noise. They attempted to assign confidence on instance label such that incorrect labels receive lower confidences. We share a similar opinion in dealing with noise data, but instance weighting via confidence itself seems to be a discarding technique rather than a correcting technique. That is, a low confidence implies an attempt to eliminate the example, while a high confidence implies keeping it. By contrast, our algorithm considers both the correctly labeled and mislabeled probability for an instance. Therefore, the loss function
$$L_{\gamma}(y,f(x))=\gamma e^{-yf(x)}+(1-\gamma)e^{yf(x)}$$
explains the attitude towards an instance: delete it with $\gamma$ and correct it by $1-\gamma$. In other words, our algorithm can be viewed as a composition technique of discarding and correcting.
For the same reason, our algorithm differs from those proposed in \cite{Gao10} and \cite{Cao12}. In their discussions, they suggested heuristic algorithms to delete or revise suspicious examples during iterations in order to improve the accuracy of AdaBoost for mislabeled data. In our algorithm, the suspicious labels are similarly revised, which is a consequence of  minimizing the modified loss function (\ref{eqn:risk}). The trusted label at each sample point is the sign of the Bayes rule and is associated with a confidence level. 

Other closely related work includes \cite{Wang13} and \cite{Zhou14}.  Both consider the same confidence level of $\bi x_i$ as $p_i=p(z_i=1|\bi x_i)$, 
whereas our approach takes advantage of the observed label $y_i$ by considering  $\gamma_i=P(z_i=y_i|\bi x_i)$.  We evaluate confidence of the observed label $y_i$, while they assess confidence of the positive label $+1$.  In \cite{Zhou14}, the initial weight $|2p_i-1|$ is very similar to our choice, but our re-weighting and classifier combination rules are different. \cite{Wang13}  has a similar combination rule as ours, but the initial weights are different.

\section{Consistency of CB-AdaBoosting}\label{sec:consistency}

In this section, we study consistency of the proposed CB-AdaBoosting method with label confidences estimated by KNN approach. Several authors have shown that the original and modified versions of AdaBoost are consistent. For example, Zhang and Yu \cite{Zhang05} considered a general ``boosting" with a step size restriction. Lugosi and Vayatis \cite{Lugosi04} proved the consistency of regularized boosting methods. Bartlett and Traskin \cite{Bartlett07} studied the stopping rule of the traditional AdaBoost that guarantees its consistency. In our algorithm, we use the exponential loss function. We just use a different empirical version of the exponential risk.  This enables us to adopt the stopping strategy used in \cite{Bartlett07} with a consistency result on the nearest neighborhood method (\cite{Stone77, Devroye94}) to show that the proposed CB-AdaBoost is Bayes-risk consistent.

%
We use notation similar to \cite{Bartlett07}. Let $(\bi X,Z)$ be a pair of random values in $ {\mathbb R}^p \times \{ -1,1 \}$ with the joint distribution $P_{X,Z}$ and the marginal distribution of $\bi X$ being $P_X$.  The training sample data ${\cal L}_n = \{(\bi x_1, y_1), ...,(\bi x_n, y_n)\}$ is available, having the same distribution as  $(\bi X,Z)$.  The mislabel problem can be treated as the case $P_{X,Z}$ being a contamination distribution. The CB-AdaBoost produces a classifier $g_n = \mbox{sign}(f_n): {\mathbb R}^p \rightarrow \{-1,1\}$ based on this sample ${\cal L}_n$. The misclassification probability  is given by 
$$ L(g_n) = P (g_n(\bi X)\neq Z|{\cal L}_n). $$ 
Our goal is to prove that $L(g_n)$  approaches the Bayes risk
$$L^* = \inf _f L(f) = \E (\min (\eta(\bi X), 1-\eta(\bi X))),$$
as $n\rightarrow \infty$, where the infimum is taken over all measurable classifiers and where $\eta(\bi X)$ is the conditional probability $\eta(\bi X)=P(Z=1|\bi X)$.

Assume that ${\cal H}$ is the set of all linear combinations of base classifiers and has a finite VC dimension. The proposed CB-AdaBoost finds a combination $f$ in ${\cal H}$ that minimizes 
$$R_{n,k_n}(f) = \frac{1}{n} \sum _{i=1}^n [\hat \gamma_i \exp(-y_i f(\bi x_i))+(1-\hat \gamma_i)\exp(y_if(\bi x_i))],$$
where $\hat\gamma_i$ is a K-NN estimator of $\gamma_i = P(Z=y_i|\bi x_i)$. That is,
$$\hat\gamma_i = \frac{1}{k_n}\sum_{j=1}^{k_n}\sum_{\bi x_j\in {\cal N}(\bi x_i)} I (y_j =y_i), $$
where ${\cal N}(\bi x_i)$ denotes the set containing $k_n$ nearest neighbors of $\bi x_i$.  
We denote 
$$\bar{R}_{n}(f) = \frac{1}{n} \sum _{i=1}^n [\gamma_i \exp(-y_i f(\bi x_i))+(1-\gamma_i) \exp(y_i f(\bi x_i))], $$
and the true exponential risk as 
$$R (f) = \E_{\bi x}\E_{z|\bi x} \exp(-Zf(\bi X))=\E\exp(-Zf(\bi X).$$
We first prove that the CB-Adaboost is consistent with the exponential risk. Then, by \cite{Bartlett06}, its 0-1 risk also approaches the Bayes risk $L^*$, since the exponential loss is classification calibrated. 

We shall denote the convex hull of ${\cal H}$ scaled by $\lambda \geq 0$ as 
$${\cal F}_{\lambda} =\{f | f = \sum_{i=1}^{n} \beta_i h_i, n\in {\cal N} \cup \{0\}, \sum_{i=1}^n \beta_i=\lambda, h_i \in {\cal H}\},$$
and the set of $t$-combinations, $t\in {\cal N}$, of functions in $\cal H$ is denoted as 
$$ {\cal F}^t = \{f | f =\sum_{i=1}^{t} \beta_i h_i , \beta_i \in \mathbb{R}, h_i \in {\cal H}\}.$$
Define the truncated function $\pi_l(\cdot)$ to be 
$$\pi _l(x) = x I(x\in [-l,l]) + l  \mbox{sign}(x),$$ 
where $I(x)$ is the indicator function.  The set of truncated functions is $\pi_l \circ {\cal F} =\{ \tilde{f} |\tilde{f} = \pi_l(f), f\in {\cal F}\}$ and the set of classifiers based on a class ${\cal F}$ is denoted by $g\circ {\cal F} = \{ \tilde{f} |\tilde{f} = g(f), f\in {\cal F}\}$. 

Based on the stopping strategy of \cite{Bartlett07} and the universal consistency of nearest neighbor function estimate of \cite{Devroye94}, we have the following proposition.\\

{\bf Proposition 4}. {\em Assume that $V = d_{VC}({\cal H}) < \infty$ and that $\cal H$ is dense in the sense of 
$\lim_{\lambda \rightarrow \infty} \inf_{f\in {\cal F}_{\lambda} }R(f) = R^*.$
Further assume $k_n \rightarrow \infty$, $k_n /n \rightarrow 0$ and $t_n =n^{1-a}$ for $a \in (0,1)$. Then the CB-AdaBoost stopped at step $t_n$ returns a sequence of classifiers almost surely satisfying $L(g(f_n)) \rightarrow L^*$.} \\

The proposition states the strong consistency of the proposed CB-AdaBoost method if it stops at $t_n = n^{1-a}$ and the size of neighbors for estimating label confidence $k_n \rightarrow \infty$ but $k_n/n \rightarrow 0$. A proof of Proposition 4 is given in Appendix.

\begin{figure*}[tbh]
\hspace{-0.4in}
\includegraphics[height=5in, width=7.8in]{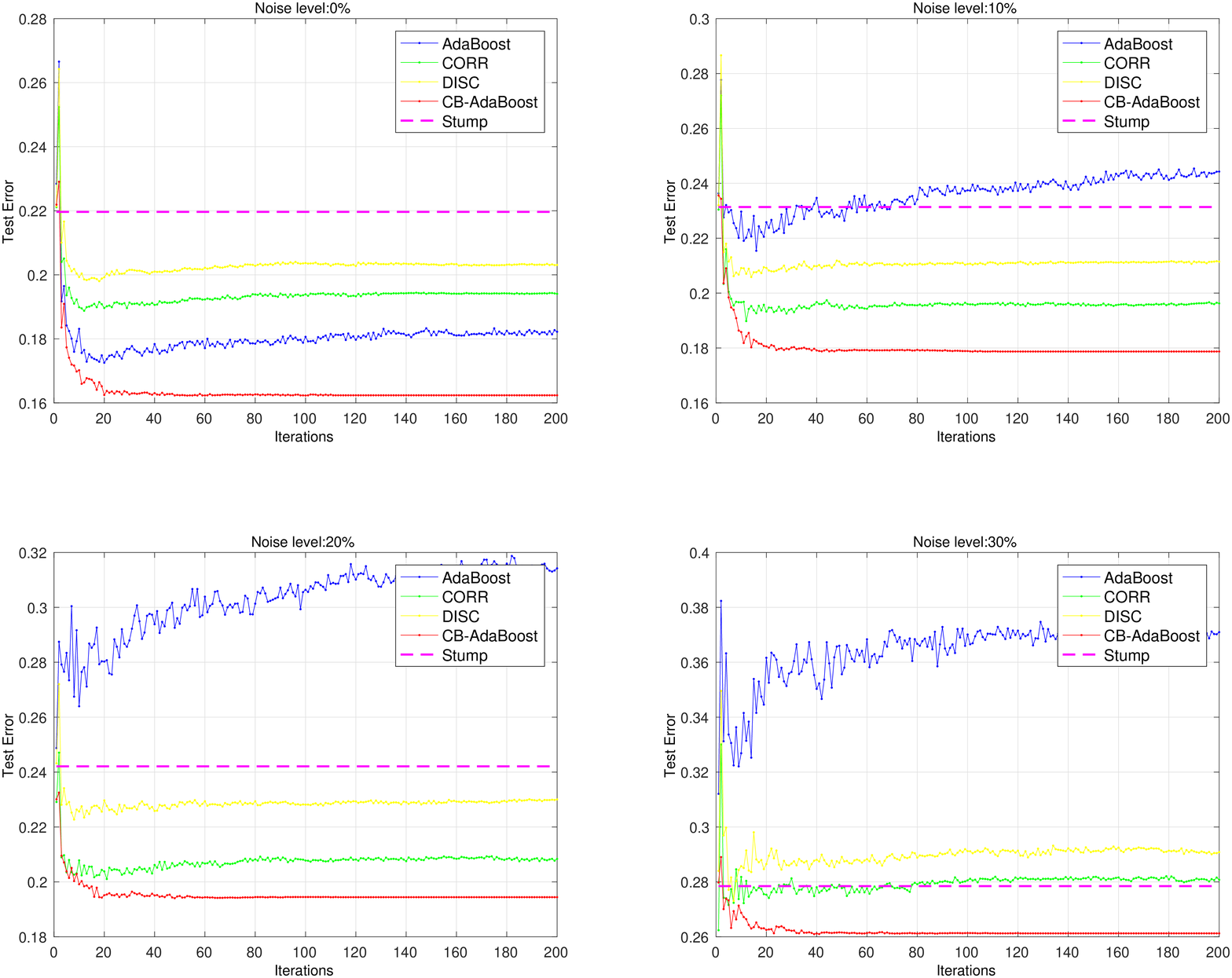}\vspace{-0.2in}
\caption{Testing errors of each method under different noise levels (0\%,10\%, 20\% and 30\%) as the number of iterations increases. }
\label{fig:Training} 
\end{figure*}

\section{Experiments}\label{sec:simulation}
To begin, we run three experiments to investigate performance of our proposed algorithm on synthetic data.  The first one examines the quality of assigned label confidence, since it has a great impact on the effectiveness of the proposed method. The second explores the advantages of the proposed algorithm over other commonly used methods in dealing with noise data.  The third experiment  demonstrates significant differences of weights between the proposed algorithm and the original AdaBoost method.  We generate random samples from two scenarios with increasing levels of label noise.  
\begin{description}
\item[Norm]  Two classes of data are sampled from bivariate normal distributions $N((0,0)^T,I)$ and $N((2,2)^T,I)$,  respectively.
\item[Sine]  Random vectors $\bi x_i=(x_{i1},x_{i2} )^T$ uniformly distributed on $[-3,3] \times [-3,3]$ are simulated, and their labels are assigned according to the conditional probability $P(z=y|\bi x_i)=e^{yg(\bi x_i)}/(e^{yg(\bi x_i)}+e^{-yg(\bi x_i)} )$, where $y\in \{1,-1\}$ and
$g(\bi x_i)=((x_{i2}-3 \sin x_{i1} )) /2$.
\end{description} 

Data sets consist of 50/500 training observations and 10000 testing instances. We introduce mislabeled data by randomly choosing training instances and reversing their labels.  

We then carry out the experiment on real data sets from the UCI repository \cite{Lichman13}. Seventeen data sets of different sizes with different numbers of input variables are used to compare performance of the proposed algorithm with some existing robust boosting methods. 

We set the number of iterations $M$ to be 200 for all ensemble classifiers. The base classifier used in the AdaBoost and CB-AdaBoost is the classification stump, the simplest one-level decision tree. 

\begin{table*}[thb]
\renewcommand{\arraystretch}{1.1}
\centering
\scriptsize
\begin{tabular}{lllcccccccr} \hline\hline
Data&Level&$n$& AdaBoost&DISC20&DISC50&DISC80&CORR20&CORR50&CORR80&CB-AdaBoost\\ \hline
Normal& 0\%&50&.1453$\pm$.0302&.1407$\pm$.0308&.1510$\pm$.0410&.1719$\pm$.0410& .1457$\pm$.0537&.1413$\pm$.0429&.1484$\pm$.0272&{\bf.1070$\pm$.0168}\\
&& 500&.0942$\pm$.0040&.0863$\pm$.0038&.0848$\pm$.0037&.0899$\pm$.0074&.0857$\pm$.0037&.0833$\pm$.0026&.0888$\pm$.0044&{\bf.0809$\pm$.0032}\\[1ex]
    
&10\%& 50&.1979$\pm$.0419&.1447$\pm$.0464&.1562$\pm$.0452&.1779$\pm$.0544&.1410$\pm$.0458&.1468$\pm$.0441&.1482$\pm$.0627&{\bf .1128$\pm$.0269}\\
&&500&.1296$\pm$.0108&.0901$\pm$.0044&.0881$\pm$.0049&.0949$\pm$.0078&.0895$\pm$.0042&.0863$\pm$.0044&.0921$\pm$.0052&{\bf.0835$\pm$.0048}\\[1ex]
    
&20\%&50&.2749$\pm$.0576&.1678$\pm$.0418&.1769$\pm$.0528&.2041$\pm$.0490&.1651$\pm$.0453&.1782$\pm$.0530&.1900$\pm$.0577&{\bf.1390$\pm$.0366}\\

&&500&.1742$\pm$.0183&.0967$\pm$.0069&.0882$\pm$.0047&.1048$\pm$.0142&.0953$\pm$.0070&.0857$\pm$.0047&.1099$\pm$.0164&{\bf.0849$\pm$.0050}\\[1ex]
     
&30\%&50&.3446$\pm$.0391&.2602$\pm$.0771&.2450$\pm$.0872&.3270$\pm$.1381&.2679$\pm$.0782&.2497$\pm$.1169&.2994$\pm$.0882&{\bf .2375$\pm$.1195}\\

&&500&.2474$\pm$.0298&.1457$\pm$.0205&.1015$\pm$.0134&.6049$\pm$.4305&.1410$\pm$.0220&{\bf.1014$\pm$.0131}&.2397$\pm$.0699&.1028$\pm$.0173\\[1ex] \hline

Sine&0\%& 50&.2305$\pm$.0221&.2271$\pm$.0272&.2358$\pm$.0508&.2420$\pm$.0281&.2306$\pm$.0316&.2307$\pm$.0330&.2402$\pm$.0228&{\bf .2139$\pm$.0188}\\
     & &500&.1934$\pm$.0074&.1850$\pm$.0071&.1859$\pm$.0090&.1871$\pm$.0076&.1851$\pm$.0073&.1861$\pm$.0083&.1891$\pm$.0087&{\bf.1834$\pm$.0067}\\[1ex]
     
&10\%&  50& .2872$\pm$.0303&.2497$\pm$.0450&.2430$\pm$.0343&.2469$\pm$.0353&.2408$\pm$.0310&.2428$\pm$.0328&.2566$\pm$.0339&{\bf .2318$\pm$.0299}\\
    & &500&.2242$\pm$.0086&.1961$\pm$.0089&.1934$\pm$.0083&.1902$\pm$.0081&.1954$\pm$.0099&.1926$\pm$.0085&.1931$\pm$.0100&{\bf.1887$\pm$.0098}\\[1ex]
    
&20\% &50&.3247$\pm$.0433&.2782$\pm$.0406&.2761$\pm$.0395&.2848$\pm$.0717&.2754$\pm$.0432&.2786$\pm$.0494&.2978$\pm$.0558&{\bf.2672$\pm$.0540}\\
     &&500&.2641$\pm$.0162&.2295$\pm$.0135&.2236$\pm$.0135&.2166$\pm$.0147&.2250$\pm$.0129&.2217$\pm$.0140&.2275$\pm$.0135&{\bf.2096$\pm$.0168}\\[1ex]

& 30\%  &50&.4017$\pm$.0545&.3349$\pm$.0711&.3433$\pm$.0822&.3448$\pm$.0793&.3338$\pm$.0709&.3320$\pm$.0766&.3497$\pm$.0687&{\bf.3258$\pm$.0811}\\
   
    &&500&.3166$\pm$.0310&.2676$\pm$.0292&.2671$\pm$.0270&.2576$\pm$.0264&.2659$\pm$.0267&.2661$\pm$.0285&.2654$\pm$.0290&{\bf.2264$\pm$.0278} \\ \hline\hline\\
\end{tabular}
\caption{Average and standard deviation of testing errors of each method under different noise levels. Discarding and correcting methods use 0.20, 0.50 and 0.80 as the confidence thresholds. The smallest errors are shown in bold. \label{testerror}}
\end{table*}

\subsection{Assessing the quality of label confidence}
It is expected that the label confidence of clean sample instances shall be high, while for mislabelled instances, the confidence should be low. In this experiment, we examine two assignment methods previously introduced in Section \ref{sec:assignment} by assessing the quality of their label confidence results.  We use the Bayesian method on the Normal data in which the noise level is known to be 0\%, 10\% and 20\%, respectively. The KNN method is used on the Sine data.  The number of nearest neighbors ($K$) used in KNN is selected from the range 3 to 15, and is set to  5 for consideration of balance between accuracy and computation efficiency. 

Table \ref{conflevel} reports the average and standard deviation of confidences on clean and mislabeled samples. The averages and standard deviations are calculated through 30 repetitions. As expected, there exists a significant separation between two types of samples on confidences. On average, clean labels  achieve a higher degree of confidence than corrupted ones. For example, under 10\% contamination of normal samples of size $n=500$, confidence for clean sample is 0.9172 compared to 0.0855 for mislabeled sample. For the small size $n=50$, the confidence difference is also significant with 0.8919 for clean sample and 0.0581 for mislabeled sample.  As the noise level increases, the difference of label confidences between clean data and mislabeled data becomes smaller. This phenomenon mentioned in \cite{Rebbapragada07} is understandable because the certainty decreases in high noise data and because assignment methods tend to be more conservative than they are in low noise data. 

\begin{figure*}[thb]
\centering
\includegraphics[width=7.5in]{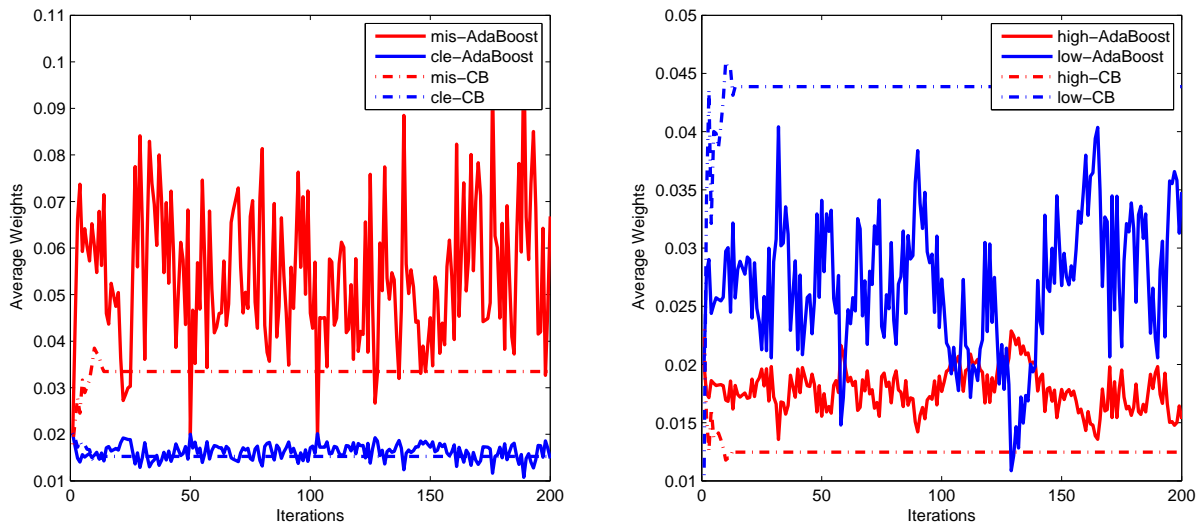}\vspace{-0.2in}
\caption{Average weights of different types of instances during the learning process in the original AdaBoost and CB-AdaBoost. The left panel is for weights of mislabelled instances and clean-labelled instances. The right panel is for weights of groups with high and low label confidence.  }
\label{fig:Weight}
\end{figure*}

\subsection{Comparisons with discarding and correcting methods}
We compare the efficiency of  the label-confidence based learning with the discarding and correcting techniques. For the latter two, a threshold on confidence is pre-specified to define suspect samples. We consider four types of classifiers:  1) AdaBoost;  2) AdaBoost working on the data with suspected samples having been discarded (DISC); 3) AdaBoost working on the original training set but suspected labels having been corrected (CORR); 4) CB-AdaBoost. We repeat the procedure 30 times and record the test errors of the four classifiers.

Fig. \ref{fig:Training} illustrates how the average test error changes as the number of iterations increases for different classifiers based on the training set of size 50.  The threshold is set to 0.5 for DISC and CORR methods.  AdaBoost greatly improves the prediction accuracy of Stump (a simple one-level decision tree) in the clean data, but its ability in boosting is limited when the training set is corrupted, especially at high noise levels where it performs even worse than a single stump. This demonstrates that AdaBoost is indeed very sensitive to noise. It also suffers from overfitting at 0\% noise level if the number of iterations becomes large. With preprocessing techniques (CORR or DISC), AdaBoost acts well at the beginning but its accuracy decreases as a large number of base learners accumulate.  Compared with the above methods, our proposed algorithm shows better performance in clean data and better robustness against noise. Moreover, it tactically avoid overfitting by ceasing the learning process at an early iteration (as early as 40).

Table \ref{testerror} provides test errors for correcting and discarding methods under different thresholds of 0.2, 0.5 and 0.8, denoted as DISC20, DISC50 and DISC80 or CORR20, CORR50 and CORR80, respectively. We now see that CB-AdaBoost's performance is superior  in 15 out of 16 cases. The only exception is the normal data under 30\% noise level for $n=500$, where CORR50 and DISC50 perform better. The advantage of CB-AdaBoost over the others is more significant for smaller sizes than for larger sizes. Neither the correcting nor discarding method at one threshold performs uniformly better than other thresholds. This makes their practice use difficult with reasonable confidence thresholds. It is worthwhile to mention that CB-AdaBoost uniformly outperforms AdaBoost even for the case without mislabels. This is because there is overlapping between the two classes and because the proposed loss function considers true risks that may help the classification achieve a better performance,  with a test error close to the theoretically minimum error, namely Bayes error. 

\begin{table*}[thb]
\centering
\begin{tabular}{lccc||lccc} \hline\hline
Datasets& Instances& Input variables& Original classes&Datasets& Instances& Input variables& Original classes\\ \hline
Breast-Cancer&683&9& 2&Wine&178&13&3\\
Wpbc& 194&33&2 &Haberman&306&3&2\\
Wdbc &569&30&2 &Vehicle&846&18&4\\
Pima&768&8&2&Banknote&1372&4&2\\
Aust&690&14&2&Cardiotocography&2126&21&3 \\
Heart&270&13&2&Waveform&5000&21&3\\
Glass&214&9&6&Urban Land Cover&	181&	147&	2\\
Seeds&210&7&3&Musk&	6598&	168	&2\\
Ecoli&336&7&8 &&&&\\
 \hline\hline
\end{tabular}
\caption{Summaries of Data sets.  \label{datasets}}
\end{table*}

\subsection{Reweighting}
This experiment illustrates re-weighting differences between the original AdaBoost and the proposed one. Fig. \ref{fig:Weight} plots how the average weights of different groups of instances change as the number of iterations increases.  First, we consider two groups: mislabeled instances and clean-labeled instances. Their mean weights are plotted in the left panel in Fig. \ref{fig:Weight}.  As the learning process continues, the mean weight of noise data in AdaBoost (mis-AdaBoost, the top red curve) rapidly rises and stays at a level much higher than that of CB-AdaBoost (mis-CB, the middle red curve). If the iterations cannot be stopped in time, the weak classifiers trained by heavily-weighted noise data become unreliable. By contrast, our proposed method does not place too much weight on noisy examples.

The right panel of Fig. \ref{fig:Weight} illustrates groups divided by the certainty degree (higher than 0.7 or not). The plot clearly demonstrates the features of the weighting rule in CB-AdaBoost. Instances with high certainty are initialized more and their average weights decline after being fully trained whereas the average weights of the others increase and remain at high values until the iteration stops. However, this adaptive ability is not present in AdaBoost.

\begin{table*}[thb]
\renewcommand{\arraystretch}{1.1}
\centering
\begin{tabular}{lcccccr} \hline\hline
Dataset&Stump&$\beta$-Boosting&MadaBoost&LogitBoost&AdaBoost&CB-AdaBoost\\ \hline\hline
 \underline{$10\%$ Noise Level}&\\ 
Breast-Cancer&.0792$\pm$.0173&.0481$\pm$.0117&.0652$\pm$.0150&{\bf.0470$\pm$.0141}&.0850$\pm$.0180&.0495$\pm$.0113\\
Wpbc&.2945$\pm$.0581&.3048$\pm$.0540&.3024$\pm$.0420&.3069$\pm$.1374&.3093$\pm$.0470&{\bf.2670$\pm$.0383}\\
Wdbc&.1008$\pm$.0148&.0786$\pm$.0190&.1098$\pm$.0209&.0719$\pm$.0170&.1043$\pm$.0228&{\bf.0589$\pm$.0166}\\
     Pima&.2845$\pm$.0327&.2696$\pm$.0197&.2845$\pm$.0210&{\bf.2424$\pm$.0156}&.3036$\pm$.0296&.2555$\pm$.0213\\
    Aust&.2663$\pm$.1537&{\bf.1478$\pm$.0173}&.1845$\pm$.0188&.1724$\pm$.1570&.2016$\pm$.0278&.1694$\pm$.0621\\
   Heart&.2998$\pm$.0477&.2842$\pm$.0836&.2649$\pm$.0403&.2183$\pm$.0535&.2719$\pm$.0363&{\bf.2015$\pm$.0288}\\
   Glass&.3318$\pm$.0513&.2673$\pm$.0466&.2717$\pm$.0522&{\bf.2417$\pm$.0476}&.2785$\pm$.0498&.2667$\pm$.0602\\
   Seeds&.3286$\pm$.0819&.1263$\pm$.0316&.1422$\pm$.0424&.1137$\pm$.0378&.1400$\pm$.0463&{\bf.1070$\pm$.0258}\\
   Ecoli&.1460$\pm$.0290&.0879$\pm$.0244&.1175$\pm$.0308&.0683$\pm$.0196&.1222$\pm$.0246&{\bf.0589$\pm$.0212}\\
    Wine&.1030$\pm$.0424&.0839$\pm$.0361&.1011$\pm$.0413&.0846$\pm$.0376&.1034$\pm$.0373&{\bf.0472$\pm$.0256}\\
Harberman&.2767$\pm$.0354&.2950$\pm$.0330&.3187$\pm$.0327&.2693$\pm$.0310&.3429$\pm$.0418&{\bf.2641$\pm$.0307}\\
 Vehicle&.2586$\pm$.0211&.0621$\pm$.0145&.0742$\pm$.0163&{\bf.0574$\pm$.0138}&.1012$\pm$.0162&.0652$\pm$.0158\\ 
Banknote&.1598$\pm$.0138&.0201$\pm$.0084&.0218$\pm$.0074&.0332$\pm$.0180&.0418$\pm$.0115&{\bf .0116$\pm$.0093}\\ 
Cardiotocography&.1611$\pm$.0199&{\bf .0780$\pm$.0079}&.0786$\pm$.0078&.0861$\pm$.0125&.1024$\pm$.0111&.0917$\pm$.0104\\
Waveform& .2322$\pm$.0114&.1147$\pm$.0064&.1138$\pm$.0047&.1289$\pm$.0105&.1350$\pm$.0083&{\bf .1114$\pm$.0059} \\ 
Urban Land Cover&	.0524$\pm$.0285&	.1535$\pm$.2326&	.0941$\pm$.0334&	.0883$\pm$.0340 & .0835$\pm$.0341&{\bf .0333$\pm$.0151}\\
Musk	&.3384$\pm$.0620&	.2577$\pm$.0297&	.2536$\pm$.0340&	.2612$\pm$.0403	&{\bf .2524$\pm$.0337}&	.2762$\pm$.0366\\
\hline

\underline{$20\%$ Noise Level}&\\
Breast-Cancer&.0986$\pm$.0241&.0589$\pm$.0163&.0875$\pm$.0175&{\bf.0543$\pm$.0169}&.1193$\pm$.0223&.0562$\pm$.0184\\
         Wpbc&{\bf.2883$\pm$.0556}&.3443$\pm$.0407&.3505$\pm$.0528&.3265$\pm$.1384&.3485$\pm$.0515&.2921$\pm$.0438\\
         Wdbc&.0987$\pm$.0189&.1280$\pm$.0247&.1860$\pm$.0261&.0878$\pm$.0217&.1801$\pm$.0305&{\bf.0743$\pm$.0216}\\
         Pima&.3084$\pm$.0519&.3023$\pm$.0251&.3283$\pm$.0295&{\bf.2618$\pm$.0170}&.3510$\pm$.0343&.2751$\pm$.0255\\
         Aust&.2946$\pm$.1741&{\bf.1648$\pm$.0215}&.2270$\pm$.0240&.1774$\pm$.0822&.2654$\pm$.0252&.1728$\pm$.0604\\
        Heart&.3136$\pm$.0825&.2721$\pm$.0467&.3094$\pm$.0512&.2896$\pm$.1506&.3156$\pm$.0530&{\bf.2222$\pm$.0356}\\
        Glass&.3380$\pm$.0558&.3150$\pm$.0520&.3174$\pm$.0545&.3171$\pm$.1366&.3044$\pm$.0515&{\bf.2897$\pm$.0528}\\
        Seeds&.3527$\pm$.0859&.2048$\pm$.0360&.2283$\pm$.0519&.1740$\pm$.0571&.2394$\pm$.0451&{\bf.1222$\pm$.0346}\\
        Ecoli&.1581$\pm$.0423&.1492$\pm$.0405&.1935$\pm$.0454&.0909$\pm$.0332&.2006$\pm$.0401&{\bf.0839$\pm$.0280}\\
         Wine&.1221$\pm$.0640&.1648$\pm$.0447&.2034$\pm$.0428&.1281$\pm$.0555&.2052$\pm$.0492&{\bf.0861$\pm$.0513}\\
     Haberman&.2972$\pm$.0854&.3440$\pm$.0480&.3564$\pm$.0456&.2756$\pm$.0347&.3793$\pm$.0362&{\bf.2706$\pm$.0242}\\
      Vehicle&.2578$\pm$.0338&.1039$\pm$.0225&.1295$\pm$.0225&.0875$\pm$.0246&.1716$\pm$.0273&{\bf.0828$\pm$.0172} \\
      Banknote&.1597$\pm$.0156&{\bf .0294$\pm$.0118}&.0460$\pm$.0173&.0587$\pm$.0184&.0841$\pm$.0216&.0376$\pm$.0145\\ 
Cardiotocography&.1653$\pm$.0197&{\bf .0993$\pm$.0128}&.1080$\pm$.0104&.1027$\pm$.087&.1537$\pm$.0164&.1046$\pm$.0148\\
Waveform& .2325$\pm$.0142&.1273$\pm$.0086&.1347$\pm$.0079&.1457$\pm$.0099&.1719$\pm$.0115&{\bf .1239$\pm$.0096} \\ 
Urban Land Cover	&.1077$\pm$.0949& .2516$\pm$ .2110&	.2084$\pm$.0471&	.1996$\pm$.0574&	.2205$\pm$.0468&	{\bf .0802$\pm$.0589}\\
Musk	&.3508$\pm$.0829&	.3064$\pm$.0381&	.3132$\pm$ .0358&	 {\bf.2957$\pm$.0461}&	.3265$\pm$ .0460&	.3234$\pm$.0308\\
\hline

\underline{$30\%$ Noise Level}&\\
Breast-Cancer&.0978$\pm$.0306&.0983$\pm$.0293&.1418$\pm$.0285&{\bf.0736$\pm$.0250}&.1835$\pm$.0406&.0805$\pm$.0302\\
         Wpbc&.3794$\pm$.1418&.4041$\pm$.0636&.3973$\pm$.0660&.3835$\pm$.0820&.4076$\pm$.0527&{\bf.3543$\pm$.0834}\\
         Wdbc&{\bf.1187$\pm$.0360}&.2339$\pm$.0388&.2919$\pm$.0408&.1262$\pm$.0375&.2953$\pm$.0383&.1209$\pm$.0403\\
         Pima&.3139$\pm$.0592&.3322$\pm$.0348&.3574$\pm$.0280&{\bf.2868$\pm$.0350}&.3930$\pm$.0337&.2924$\pm$.0340\\
         Aust&.2947$\pm$.1552&.2479$\pm$.1500&.2989$\pm$.0308&.2310$\pm$.1642&.3326$\pm$.0357&{\bf.2014$\pm$.0704}\\
        Heart&.3133$\pm$.0685&.3637$\pm$.1473&.3630$\pm$.0536&.2630$\pm$.0449&.3723$\pm$.0538&{\bf.2477$\pm$.0422}\\
        Glass&.3910$\pm$.0921&.3679$\pm$.0619&.3757$\pm$.0554&.3617$\pm$.0677&.3816$\pm$.0613&{\bf.3542$\pm$.0788}\\
        Seeds&.3622$\pm$.0839&.2876$\pm$.0683&.3321$\pm$.0629&.2559$\pm$.0862&.3371$\pm$.0572&{\bf.2029$\pm$.0632}\\
        Ecoli&.2022$\pm$.0689&.2351$\pm$.0528&.2796$\pm$.0646&.1512$\pm$.0629&.3022$\pm$.0527&{\bf.1359$\pm$.0472}\\
         Wine&.2307$\pm$.1092&.2625$\pm$.0565&.2884$\pm$.0467&.2015$\pm$.0714&.2805$\pm$.0633&{\bf.1528$\pm$.0758}\\
     Haberman&.3266$\pm$.0990&.3752$\pm$.0498&.3983$\pm$.0475&{\bf.3089$\pm$.0591}&.4157$\pm$.0512&.3205$\pm$.0767\\
      Vehicle&.2610$\pm$.0260&.1829$\pm$.0308&.2212$\pm$.0274&.1516$\pm$.0574&.2644$\pm$.0334&{\bf.1357$\pm$.0420}\\
      Banknote&.1678$\pm$.0189&.0825$\pm$.0238&.1152$\pm$.0238&.0921$\pm$.0390&.1657$\pm$.0273&{\bf .0720$\pm$.0221}\\ 
Cardiotocography&.1837$\pm$.0317&.1388$\pm$.0192&.1614$\pm$.0219&.1272$\pm$.0235&.2326$\pm$.0254&.1218$\pm$.0237\\
Waveform& .2433$\pm$.0221&.1561$\pm$.0110&.1723$\pm$.0111&.1639$\pm$.0143&.2287$\pm$.0143&{\bf .1410$\pm$.0106} \\
Urban Land Cover&	.1758$\pm$.1046&	.2868$\pm$.0513	&.2956$\pm$.0705	&.2674$\pm$.0565	&.2938$\pm$.0682	&{\bf .1524$\pm$.0746}\\
Musk &.4214$\pm$.0710&.3633$\pm$.0316&	.3800$\pm$.0471&	.3639$\pm$.1274	&.3821$\pm$.0295	&{\bf .3472$\pm$.0464}\\
 \hline\hline\\
\end{tabular}
\caption{Average and standard deviation of testing errors of each classifier. The boldface represents the smallest testing error. \label{real}}
\end{table*}

\begin{table*}[thb]
\renewcommand{\arraystretch}{1.3}
\centering
\begin{tabular}{lllllr} \hline\hline
Noise level& Stump&$\beta$-Boosting& MadaBoost&LogitBoost&AdaBoost\\ \hline
10\%&17/17***&12/17*&15/17***&11/17&16/17***\\
20\% & 16/17***&13/17**&16/17***&13/17**&17/17***\\
30\% & 16/17***&17/17***&17/17***&14/17***&17/17***\\
\hline\hline\\
\end{tabular}
\caption{The relative frequencies that CB-AdaBoost wins over 17 data sets in pairwise comparisons with other algorithms.``***", ``**" and ``*" are used if CB-AdaBoost is statistically better with 1\%, 5\% and 10\% significance level, respectively. \label{better} }
\end{table*}

\subsection{Real data sets}
In addition, we conducted experiments on 17 real data sets available from the UCI repository \cite{Lichman13}. Since we focus on the two-class problem, the classes of several multi-class data sets are combined into 2 classes.  If the class variable is nominal, class 1 is treated as the positive class, and the remaining classes are treated as the negative class.  If the class variable is ordinal, we merge the classes with similar properties. For example,  in the Cadiotocography data, ``Suspect" class and "Pathologic" class are combined as the positive class and "Normal" as the negative class.  For the Urban Land Cover dataset, we combine the training and test set, and any instances with missing values are removed.  Table \ref{datasets} summarizes the main characteristics of all data sets. For each data set, half of the instances are randomly selected as the training set and the remaining are used for testing.  10\%, 20\% and 30\% mislabels are introduced in the training data by randomly choosing training instances and reversing their labels.  For comparison, we consider another boosting method known as LogitBoost, in addition to the two modified AdaBoost algorithms known as MadaBoost \cite{Domingo00} and $\beta$-Boosting \cite{Zhang08b}~\footnote{The original paper did not name the method. We name it $\beta$-Boosting after the $\beta$ parameter added to the algorithm, as suggested by a reviewer.},  all of which are robust against noise data. The procedure is repeated 30 times, and we take the average of the 30 test errors for each classifier as a measurement of its performance.

According to Table \ref{real}, CB-AdaBoost performs better than the original AdaBoost for all cases except for one which is the case of Musk under 10\% noise level.  It also greatly improves the accuracy of the stump (i.e., the base classifier). $\beta$-Boosting, MadaBoost and LogitBoost methods show robustness to mislabeled data. They outperform AdaBoost for most cases especially which LogitBoost achieves a lower test error than the other two.  However, like AdaBoost, they suffer from overfitting because they cannot stop iterations due to their weight distributions. This problem is overcome by CB-AdaBoost, and as a result, the win-lose numbers of the proposed algorithm when compared to the robust three algorithms are 42-9, 48-3 and 38-13 respectively. We conducted the sign test based on counts of wins, losses and ties \cite{Demsar06} in order to quantify the significance of the proposed method. Table \ref{better} lists the frequency and significance level that CB-AdaBoost wins each of other algorithms on 17 data sets at each noise level. This  demonstrates the effectiveness and advantages of CB-AdaBoost in handling mislabeled data.

\section{Conclusion}\label{sec:conclusion}
In this paper, we have provided a label-confidence based boosting method that is sufficiently immune to the label noise and overfitting problems.  With the assignment of confidence, our proposed algorithm distinguishes between clean and contaminated instances. In addition, the values of confidence on instances represent different levels of judgments on their label reliability.  Under the guidance of confident instances, CB-AdaBoost is able to minimize the loss function over the training set under the conditional risk function. Moreover, in CB-AdaBoost, explicit solutions for weak learners and their coefficients on each stage can be easily obtained and applied practically. In comparisons with some common noise handling techniques and other robust algorithms, CB-AdaBoost does a better job of tackling problems of class overlapping and mislabelling.
 
The proposed method has some limitations.
\begin{itemize}
\item  The computational complexity of the proposed CB-AdaBoost is $O(n^2d)$, where $n$ is the sample size and $d$ is dimension. This is because we need to compute or estimate label confidence of each instance,  and the KNN method for label confidence evaluation has the computation complexity $O(n^2d)$. The remaining process of the CB-AdaBoost is $O(n^{2-a}d)$ with $a \in (0,1)$. Collectively, this yields an overall computational complexity $O(n^2d)$,   which may be prohibitive for large-scale applications. 
\item As currently formulated, the proposed method cannot directly handle categorical or symbolic features. A similarity metric on those type of features needs to be introduced to define ``neighbors" for label confidence assignment. 
\end{itemize}
Continuation of this work could take several directions.  
\begin{itemize}
\item A general framework of optimization strategy based on the conditional risk deserves a deeper understanding and further development.  
\item In the current work, KNN is used to estimate the confidence of each instance. Theoretically, the number of neighbors shall go to infinity with a speed slower than the sample size to ensure strong consistency of KNN estimator.  In practice,  however,  a small number of neighbors seem to be sufficient. Perhaps a proof of consistence exists without the conditions of $k_n$.  It will be interesting to study the impact of parameter $k$ and discuss a proper selection on the number of neighborhoods in practice.  For example, the cross-validation method for choosing $k$ deserves further investigation. In fact, the problem of how to design a good criterion for confidence assignment is still open. Other methods are needed to produce high quality confidences, especially when categorical features are involved. 
 \item CB-AdaBoost outperforms the AdaBoost for class overlapping problems, thus it is promising to extend CB-AdaBoost for multiple class classification problems and other applications such as image or object recognition. 
\end{itemize}

\section{Appendix}\label{sec:appendix}
{\bf Proof of Proposition 4}.   Let $\{\bar{f}_n\}_{n=1}^\infty$ be a sequence of reference functions such that  $R(\bar{f_n})\rightarrow R^*$. We shall prove that there exist non-negative sequences $t_n \rightarrow \infty$, $\xi_n \rightarrow \infty$, $k_n \rightarrow \infty$, and $k_n/n \rightarrow 0$ such that the following conditions are satisfied. 
\begin{description}
\item [Uniform Convergence of $t_n$-combinations]  
\begin{align} 
&\sup_{f\in \pi_{\xi_n}\circ {\cal F}^{t_n}} |R(f) -\bar{R}_n(f)| \overset{a.s.}{\longrightarrow} 0; \label{uniform1}
 \end{align}
\item [Empirical convergence for the sequence $\{\bar f_n\}$]  
\begin{align}
&\bar R_{n}(\bar f_n) - R(\bar f_n) \overset{a.s.}{\longrightarrow} 0; \label{emconv1}
\end{align}
\item [Convergence of the KNN estimates]
\begin{align} 
&R_{n,k_n}(\bar f_n) - \bar R_n(\bar f_n) \overset{a.s.}{\longrightarrow} 0; \label{knnconv1}
\end{align}
\item [Algorithm convergence of $t_n$-combinations]
\begin{align}
&R_{n,k_n} (f_{t_n}) -R_{n,k_n} (\bar f_n)  \overset{a.s}{\longrightarrow} 0. \label{alconv}
\end{align}
\end{description}

Since $\bar{R}_n(f)$ is an empirical exponential risk, a proof of  (\ref{uniform1}) follows exactly the same lines of Lemma 4 in \cite{Bartlett07} with the Lipschitz constant $L_{\xi} = (e^{\xi}-e^{-\xi})/(2\xi)$ and $M_{\xi} = e^{\xi}$. Then for any $\delta>0$, with probability at least $1-\delta$,
\begin{align}
&\sup_{f \in \pi_{\xi}\circ{\cal F}^t}|R(f)-\bar{R}_n(f)| \nonumber\\
&\leq c \xi L_{\xi} \sqrt{\frac{(V+1)(t+1)\log_2[2(t+1)/\ln 2]}{n}}+M_{\xi}\sqrt{\frac{1/\delta}{2n}}, \label{upperbound}
\end{align}
where $V=d_{VC}({\cal H})$ and $c=24\int_0^1 \sqrt{\ln \frac{8e}{\epsilon^2}}d\epsilon$. We can take $t=n^{1-a}$ and $\xi = \kappa\ln n$ with $\kappa>0$, $a \in (0,1)$ and $2\kappa-a<0$ so that the right side of the inequality (\ref{upperbound}) converges to 0,  and in the mean time $\sum_{n=1}^{\infty} \delta_n< \infty$. Hence an application of the Borel-Cantelli lemma ensures the almost surely convergence of  (\ref{uniform1}).

Applying Theorem 8 of \cite{Bartlett07}, we have the result of (\ref{alconv}), in which the reference sequence $\bar f_n \in {\cal F}_{\lambda_n}$ with $\lambda_n = \kappa_1 \ln n$ where $\kappa_1 \in (0,1/2)$. 

(\ref{emconv1})  can be proved by Hoeffding's inequality if the range of $\bar f_n$ is restricted to the interval $[-\lambda_n,\lambda_n]$. That is, 
$$P(\bar R_n(\bar f_n) -R(\bar f_n) \geq \epsilon_n) \leq \exp(-2n\epsilon_n^2/M_{\lambda_n}^2):=\delta_n$$ 
where $M_{\lambda_n} = e^{\lambda_n}-e^{-\lambda_n}$. Let $\lambda_n = \kappa_1 \ln n$ with $\kappa_1 \in (0,1/2)$. Letting $\epsilon_n \rightarrow 0$, we still have $\sum_{n=1}^\infty \delta_n < \infty$,  and hence convergence in probability 1 of (\ref{emconv1}) holds. 

By the result of Theorem 1 in \cite{Devroye94}, for each KNN estimate $\hat{\gamma}_i$ with $k_n\rightarrow \infty$ and $k_n/n \rightarrow 0$, we have
$$P(2|\hat{\gamma}_i-\gamma_i|>\epsilon_n) \leq \exp [-n \epsilon_n^2/(8N_p^2)],$$ 
where the constant $N_p$ is the minimal number of cones centered at the origin of angle $\pi/6$ that cover $\mathbb{R}^p$. Then with the restriction of $\bar f_n$ in $[-\lambda_n,\lambda_n]$, we have 
\begin{align*}
& P(|R_{n,k_n}(\bar f_n)-\bar R_n (\bar f_n)| > \epsilon_n) \\
&< \exp[-n \epsilon_n^2/(2 M_{\lambda_n}^2 N_d^2)+\ln n] :=\delta_n
\end{align*}
Again, a choice of $\lambda_n = \kappa_1 \ln n$ with $\kappa_1 \in (0,1/2)$ guarantees $\sum \delta_n <\infty$ when $\epsilon_n =o(1)$, and hence (\ref{knnconv1}) holds. 

Now we are ready to prove Proposition 4. 
For almost every outcome $\omega$ on the probability space, we can define sequences $\epsilon_{n,i}(\omega) \rightarrow 0$ for $i=1,...,5$ so that for almost all $\omega$ the following inequalities are true. 
\begin{align}
&R(\pi_{\xi_n}(f_{t_n})) \nonumber\\
&\leq  \bar{R}_n(\pi_{\xi_n}(f_{t_n}))+\epsilon_{n,1}(\omega)\;\;\;\;\;\;\;\mbox{by (\ref{uniform1})}\nonumber\\
&\leq R_{n,k_n}(\pi_{\xi_n}(f_{t_n}))+\epsilon^*_{n,2}(\omega) \label{knn}\\
&\leq R_{n,k_n}(f_{t_n})+e^{-\xi_n}+\epsilon^*_{n,2}(\omega) \label{xi}\\
&\leq R_{n,k_n}(\bar{f}_n)+e^{-\xi_n}+\epsilon^*_{n,3}(\omega) \;\;\;\;\;\;\;\mbox{by (\ref{alconv})}\nonumber\\
&\leq \bar{R}_{n}(\bar{f}_n)+e^{-\xi_n}+\epsilon^*_{n,4}(\omega)\;\;\;\;\;\;\;\mbox{by (\ref{knnconv1})}\nonumber\\
&\leq R(\bar{f}_n)+e^{-\xi_n}+\epsilon^*_{n,5}(\omega)\;\;\;\;\;\;\;\mbox{by (\ref{emconv1})} \label{totalerr}
\end{align}
where $\epsilon^*_{n,k}(\omega)=\sum_{j=1}^k\epsilon_{n,j}(\omega)$. Inequality (\ref{knn}) follows similarly as (\ref{knnconv1}) with $\xi _n =\kappa_1 \ln n $,  where $\kappa_1 \in (0,1/2)$. Inequality (\ref{xi}) follows from the facts that $e^{\pi_{\xi_n}(x)}<e^x+e^{-\xi_n}$ and $e^{-\pi_{\xi_n}(x)}<e^{-x}+e^{-\xi_n}$. Then with $t_n = n^{1-a}$, $\xi_n =\kappa  \ln n$ ($a>0, \kappa>0, 2\kappa<a$) and (\ref{totalerr}), by choice of the sequence $\{\bar{f}_n\} \in {\cal F}_{\lambda_n}$ with $\lambda_n =\kappa_1\log n, \kappa_1 \in (0,1/2)$, we have $R(\bar{f}_n)\rightarrow R^*$ and $R(\pi_{\xi_n}(f_{t_n}))\rightarrow R^*$ a.s. 

By Theorem 3 of \cite{Bartlett06},  $L(g(\pi_{\xi_n}(f_{t_n})))\overset{a.s}{\longrightarrow} L^*$. Since for $\xi_n >0$ we have $g(\pi_{\xi_n}(f_{t_n}))=g(f_{t_n})$, it follows that 
$$L(g(f_{t_n})) \overset{a.s}{\longrightarrow} L^*.$$
Hence, the proposed CB AdaBoosting procedure is consistent if stopped after $t_n$ steps.

\ifCLASSOPTIONcompsoc
  \section*{Acknowledgments}
\else
  \section*{Acknowledgment}
\fi

Zhi Xiao and Bo Zhong are supported by the China National Science Foundation (NSF) under Grant No. 71171209. The authors would like to thank Yixin Chen for discussing the conditional risk.



%

\end{document}